# INDEPENDENCE AND BAYESIAN UPDATING METHODS

Rodney W. Johnson
Code 7591
Naval Research Laboratory
Washington, D.C. 20375-5000

June 14, 1985

## ABSTRACT

Duda, Hart, and Nilsson [1] have set forth a method for rule-based inference systems to use in updating the probabilities of hypotheses on the basis of multiple items of new evidence. Pednault, Zucker, and Muresan [2] claimed to give conditions under which independence assumptions made by Duda *et al.* preclude updating—that is, prevent the evidence from altering the probabilities of the hypotheses. Glymour [3] refutes Pednault *et al.*'s claim with a counterexample of a rather special form (one item of evidence is incompatible with all but one of the hypotheses); he raises, but leaves open, the question whether their result would be true with an added assumption to rule out such special cases. We show that their result does not hold even with the added assumption, but that it can nevertheless be largely salvaged. Namely, under the conditions assumed by Pednault *et al.*, at most one of the items of evidence can alter the probability of any given hypothesis; thus, although updating is possible, multiple updating for any of the hypotheses is precluded.

## BACKGROUND

Duda, Hart, and Nilsson [1] consider the problem of updating the probability of a hypothesis $H$ with prior probability $P(H)$ when new evidence is obtained in the form of propositions $E_j$ for which the conditional probabilities $P(E_j \mid H)$ and $P(E_j \mid \overline{H})$ are known. They assume that the $E_j$ are conditionally independent, both on condition $H$ and on condition $\overline{H}$, so that

$$P(E_1 \cdots E_m \mid H) = \prod_{j=1}^{m} P(E_j \mid H)$$

$$P(E_1 \cdots E_m \mid \overline{H}) = \prod_{j=1}^{m} P(E_j \mid \overline{H}).$$

They can then write an updating formula for the odds on $H$ in terms of a product of likelihood ratios:

$$\frac{P(H \mid E_1 \cdots E_m)}{P(\overline{H} \mid E_1 \cdots E_m)} = \frac{P(H)}{P(\overline{H})} \prod_{j=1}^{m} \frac{P(E_j \mid H)}{P(E_j \mid \overline{H})}.$$

Pednault, Zucker, and Muresan [2], in analyzing this updating scheme, considered the consequences of imposing the independence assumptions for each hypothesis $H_i$ of a jointly exhaustive, mutually exclusive set. They [2] and other writers (see [3] and other references therein) agree that the assumptions are unreasonably strong, but there has been some confusion over the exact extent of the undesirable consequences. Pednault *et al.* [2] concluded that if there were at least three hypotheses, then no updating could take place—that the assumptions are too strong to be satisfied unless

$$P(E_j \mid H_i) = P(E_j \mid \overline{H}_i) = P(E_j)$$

holds for all $i$ and $j$, and consequently

$$P(H_i \mid E_1 \cdots E_m) = P(H_i).$$

However, Glymour [3] gives a counterexample to their conclusion—three jointly exhaustive, mutually exclusive hypotheses $H_i$ and two evidence propositions $E_j$ that satisfy the independence assumptions but allow updating to occur. He points out that Pednault *et al.* had relied on an erroneous result claimed by Hussain [4], also refuted by his counterexample. Glymour notes that the evidence proposition $E_2$ of his counterexample has the special property that $P(E_2 \mid H_2) = P(E_2 \mid H_3) = 0$, so that $E_2$ determines a posterior probability of 1 for one hypothesis, $H_1$, and 0 for the rest. He raises, and leaves open, the question whether Pednault *et al.*'s result would be true with the additional requirement that for all $i$,

$$P(H_i \mid E_1 \cdots E_m) \neq 0. \qquad (1)$$

In the next section we answer that question by giving a counterexample that satisfies (1).

Glymour's counterexample has another special property, one that is sufficient to make it a valid counterexample: $P(E_1 \mid H_i) = P(E_1 \mid \overline{H}_i) = P(E_1)$, so that $E_1$ produces no updating; only $E_2$ produces updating. We give a second counterexample that lacks this second special property. However, we can show that Pednault et al.'s assumptions imply that for every $i$ there is at most one value of $j$ for which

$$P(E_j \mid H_i) = P(E_j \mid \overline{H}_i) = P(E_j)$$

does not hold. We show this in the section following the counterexamples. It follows that for each hypothesis $H_i$, there is at most one evidence proposition $E_j$ that produces updating of the probability of $H_i$; there is no hypothesis for which multiple updating is possible.

We conclude this section by stating the assumptions used by Pednault et al. The hypotheses are $H_1, \ldots, H_n$; it is assumed that

$$n > 2 \qquad (2)$$

and that the $H_i$ are jointly exhaustive

$$\sum_{i=1}^{n} P(H_i) = 1 \qquad (3)$$

and mutually exclusive

$$P(H_i H_j) = 0 \qquad (i \neq j). \qquad (4)$$

The evidence propositions are $E_1, \ldots, E_m$. Since a subset of an independent set is independent, we write the independence assumptions as

$$P(E_{j_1} \cdots E_{j_k} \mid H_i) = \prod_{j \in J} P(E_j \mid H_i), \qquad (5)$$

$$P(E_{j_1} \cdots E_{j_k} \mid \overline{H}_i) = \prod_{j \in J} P(E_j \mid \overline{H}_i). \qquad (6)$$

for every subset $J = \{j_1, \ldots, j_k\}$ of the indices $1, \ldots, m$.

## COUNTEREXAMPLES

For comparison, here is Glymour's counterexample.

| $H_i$: | $H_1$ | $H_2$ | $H_3$ |
|---|---|---|---|
| $P(E_1 E_2 H_i)$: | 1/6 | 0 | 0 |
| $P(E_1 \overline{E}_2 H_i)$: | 0 | 1/6 | 1/6 |
| $P(\overline{E}_1 E_2 H_i)$: | 1/6 | 0 | 0 |
| $P(\overline{E}_1 \overline{E}_2 H_i)$: | 0 | 1/6 | 1/6 |

It is straightforward to verify that

$$\begin{aligned} P(E_2 \mid H_1) &= 1, \\ P(E_2 \mid H_2) &= P(E_2 \mid H_3) = 0, \end{aligned} \qquad (7)$$

as noted by Glymour, and that

$$P(E_1 \mid H_i) = P(E_1 \mid \overline{H}_i) = P(E_1) \qquad (8)$$

for each $i$.

To answer the question raised by Glymour, we modify the example so that (7) no longer holds. We retain the same values for $P(H_i)$ and $P(E_1 \mid H_i)$, choose new, nonzero values for $P(E_2 \mid H_i)$ (say 1/2, 1/3, 1/6 for $i = 1, 2, 3$), and define the remaining relevant conditional probabilities and probabilities with the help of (5). Here is the result.

| $H_i$: | $H_1$ | $H_2$ | $H_3$ |
|---|---|---|---|
| $P(E_1 E_2 H_i)$: | 1/12 | 1/18 | 1/36 |
| $P(E_1 \overline{E}_2 H_i)$: | 1/12 | 1/9 | 5/36 |
| $P(\overline{E}_1 E_2 H_i)$: | 1/12 | 1/18 | 1/36 |
| $P(\overline{E}_1 \overline{E}_2 H_i)$: | 1/12 | 1/9 | 5/36 |

Assumptions (2)–(6) can be verified. In fact, we can show that as long as $E_1$ satisfies (8), we can choose $P(E_2 \mid H_i)$ arbitrarily, and the procedure we have just used will lead to a probability distribution that satisfies (2)–(6).

Now (8) implies that $E_1$ is irrelevant for inference about the hypotheses. Only $E_2$ produces updating—multiple updating does not occur. But by going to four hypotheses, we can dispense with (8) and obtain a counterexample such that $E_1$ and $E_2$ can both produce updating.

| $H_i$: | $H_1$ | $H_2$ | $H_3$ | $H_4$ |
|---|---|---|---|---|
| $P(E_1 E_2 H_i)$: | 1/24 | 1/12 | 1/24 | 1/12 |
| $P(E_1 \overline{E}_2 H_i)$: | 1/24 | 1/12 | 1/12 | 1/24 |
| $P(\overline{E}_1 E_2 H_i)$: | 1/12 | 1/24 | 1/24 | 1/12 |
| $P(\overline{E}_1 \overline{E}_2 H_i)$: | 1/12 | 1/24 | 1/12 | 1/24 |

Again (2)–(6) can be verified. Furthermore $E_1$ and $E_2$ can both produce updating since we have, for example, $P(E_1 \mid H_1) \neq P(E_1)$ and $P(E_2 \mid H_3) \neq P(E_2)$. However, we have

$$P(E_1 \mid H_i) = P(E_1 \mid \overline{H}_i) = P(E_1)$$
$$(i = 3, 4),$$
$$P(E_2 \mid H_i) = P(E_2 \mid \overline{H}_i) = P(E_2)$$
$$(i = 1, 2).$$

Thus only $E_1$ can update the probability of $H_1$ or $H_2$, and only $E_2$ can update the probability of $H_3$ or $H_4$; for no hypothesis is multiple updating possible. This illustrates the general case, as we show in the next section.

## IMPOSSIBILITY OF MULTIPLE UPDATING

*Theorem.* If the assumptions (2)–(6) hold, then for every $H_i$ there is at most one $E_j$ that produces updating for $H_i$.

*Proof.* No updating for $H_i$ is possible if either $P(H_i)$ or $P(\bar{H}_i)$ is 0; therefore we may assume that both are nonzero. First consider the case $m = 2$, where the evidence propositions are just $E_1$ and $E_2$. We follow Pednault *et al.* (see [2], equations (6)–(9)) in deriving

$$P(E_1)P(E_2) - P(E_1)P(E_2 H_i) - P(E_1 H_i)P(E_2)$$
$$= P(E_1 E_2)[1 - P(H_i)] - P(E_1 E_2 H_i) \quad (9)$$

and summing over $i$ to obtain

$$(n-2)P(E_1)P(E_2) = (n-2)P(E_1 E_2),$$

or

$$P(E_1)P(E_2) = P(E_1 E_2). \quad (10)$$

Using (9) and (10) with the help of (5), we obtain

$$P(E_1)P(E_2 \mid H_i)P(H_i)$$
$$\quad + P(E_1 \mid H_i)P(E_2)P(H_i)$$
$$= P(E_1)P(E_2)P(H_i)$$
$$\quad + P(E_1 \mid H_i)P(E_2 \mid H_i)P(H_i)$$

from which it follows that

$$[P(E_1) - P(E_1 \mid H_i)][P(E_2) - P(E_2 \mid H_i)] = 0.$$

One of the bracketed factors vanishes. If $P(E_j \mid H_i) = P(E_j)$, then in fact

$$P(E_j \mid H_i) = P(E_j \mid \bar{H}_i) = P(E_j);$$

this therefore holds either with $j = 1$ or with $j = 2$. Consequently $E_1$ and $E_2$ do not both produce updating for $H_i$. Thus we have proved the theorem for the case $m = 2$. But in the general case, if one of the evidence propositions, say $E_j$, produces updating for $H_i$, the result for $m = 2$ implies that no other evidence proposition $E_k$ produces updating for $H_i$.

## DISCUSSION

Assumptions (2)–(6) lead to unreasonably severe restrictions on the possibility of probabilistic updating. If we wish to make inferences about more than two jointly exhaustive, mutually exclusive hypotheses, and if we wish to allow more than one piece of evidence to bear on one hypothesis, then we must eliminate either (5) or (6). Duda *et al.* made the assumptions (5) and (6) in the context of a single pair of hypotheses $H, \bar{H}$. It is clearly a mistake to carry them both over into the context of several exhaustive, mutually exclusive hypotheses.

Cases where (5) is justified (at least as an approximation) are quite common, but (6) is much less plausible. Consider, for example, a physical quantity $x$ that can take any of a considerable range of numerical values $v_i$. Let $y$ and $z$ be measurements of $x$ made with instruments subject to independent errors; that is, suppose $y - x$ and $z - x$ are independent random variables but are fairly small with high probability. Then $y - z$ is small with high probability, and so $y$ and $z$ are highly dependent; but on condition of a given value of $x$, say $x = v_i$, the conditional distributions of $y$ and $z$ are independent. Define $H_i$ to be $x = v_i$. Then $y$ and $z$ are independent on condition $H_i$ for any $i$. However, on condition $\bar{H}_i$, we expect them to be dependent, for the same reason that their unconditional distributions are dependent. In that case, we can take $E_1$ and $E_2$ to be propositions about $y$ and $z$, respectively, and it is easy to choose $E_1$ and $E_2$ so that they are conditionally independent on condition $H_i$ but not on condition $\bar{H}_i$.

We must conclude that for inference about jointly exhaustive, mutually exclusive hypotheses, updating schemes based on the independence assumption (5) alone may be useful, but schemes based on both (5) and (6) are too restrictive to be useful.